# Development of a mobile robot assistant for wind turbines manufacturing

Ali Ahmad Malik

School of Engineering & Computer Science
Oakland University, Michigan, United States
*Email: aliahmadmalik@oakland.edu*

**Pre-print**

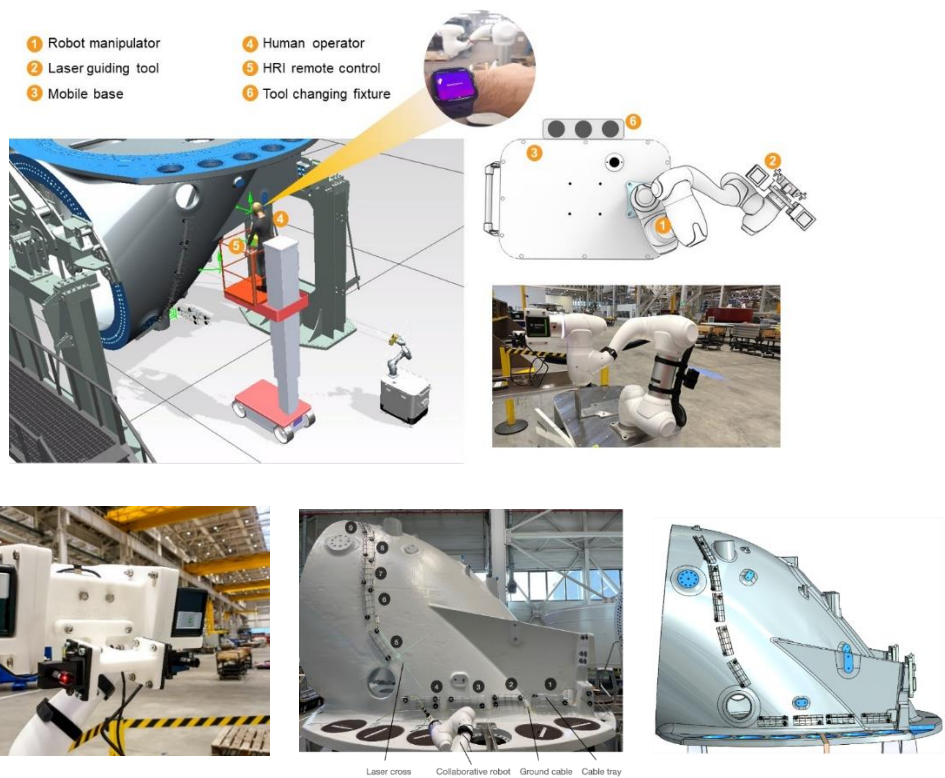

## Highlights

1. The thrust for increased rating capacity of wind turbines has resulted into larger generators, longer blades, and taller towers.
2. Due to frequent design changes and variety of tasks involved, conventional automation is not possible making it a labor-intensive activity.
3. The article presents a systematic framework and development of a mobile robot assistant to assist operators in nacelle assembly.
4. Digital twin for system verification, human-robot interaction technologies and future research directions are presented.

# Development of a mobile robot assistant for wind turbines manufacturing

Ali Ahmad Malik
School of Engineering & Computer Science
Oakland University, Michigan, United States
*Email: aliahmadmalik@oakland.edu*

**Abstract**

The thrust for increased rating capacity of wind turbines has resulted into larger generators, longer blades, and taller towers. Presently, up to 16 MW wind turbines are being offered by wind turbines manufacturers which is nearly a 60 percent increase in the design capacity over the last five years. Manufacturing of these turbines involves assembling of gigantic sized components. Due to the frequent design changes and the variety of tasks involved, conventional automation is not possible making it a labor-intensive activity. However, the handling and assembling of large components are challenging the human capabilities. The article proposes the use of mobile robotic assistants for partial automation of wind turbines manufacturing. The robotic assistant can result into reducing production costs, and better work conditions. The article presents development of a robot assistant for human operators to effectively perform assembly of wind turbines. The case is from a world's leading wind turbines manufacturer. The developed system is also applicable to other cases of large component manufacturing involving intensive manual effort.

## 1. Introduction

Wind power technology is an environmentally sound source of energy. The present global wind capacity is 743 GW (Lee & Zhao, 2021) which has increased from 24 GW in 2001 (Wiser, Millstein, et al., 2021). Wind contributes to 6% of global energy needs and is the second biggest source of green energy after hydro power ("Glob. Energy Rev. 2019," 2021) while the year over year increase is exhibited as 55% (Lee & Zhao, 2021). During an effective operational life, a wind turbine can payback the energy used in its manufacturing 23 to 57 times (Jensen, 2019).

Wind turbine manufacturers are facing a fierce market competition to offer turbines with high rated capacity. The rationale for the design capacity race is the rotor swept area is directly proportional to the energy produced. A wind turbine with high rated capacity progressively lowers levelized cost of energy (LCOE) (Wiser, Millstein, et al., 2021)(Wiser, Rand, et al., 2021). OEMs are offering wind turbines of up to 15 MW which is 8 times of what was being offered in 1998. The rotor diameter of 15 MW wind turbine is 240m which is three time the size of 2 MW wind turbine (GE Renewable Energy, 2018) (NREL, 2020) (see Figure 1).

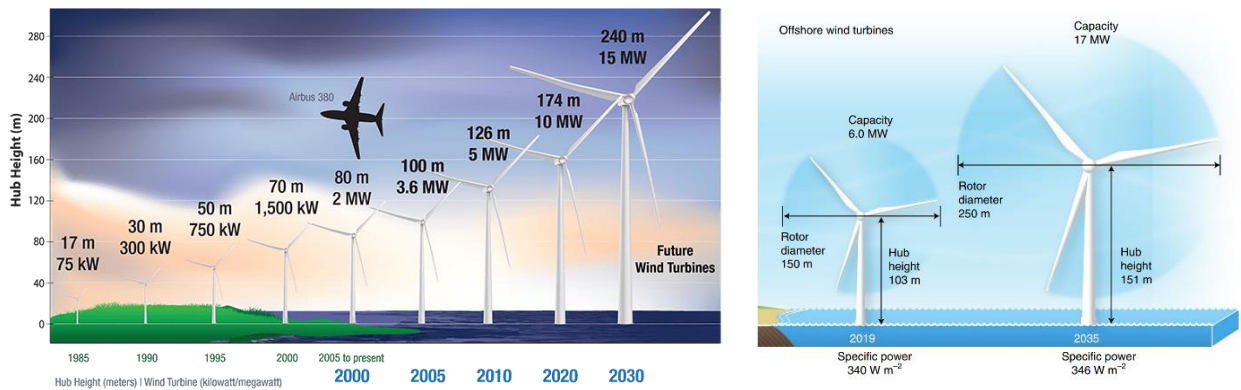

Figure 1: (a) Hub height and rated capacity of wind turbines (left) (NREL, 2020); (b) The expected size increase in offshore wind turbines (right) (Wiser, Rand, et al., 2021).

Workforce shortage is another challenge that wind turbines manufacturers have to deal with. A study on wind turbines manufacturing value chains (Surana et al., 2020) identified nacelle as a component with high degree of complexity. High complexity components of wind turbines are produced in Europe, United States and Japan (see Figure 2). Many of these countries are facing three major issues in relation to workforce availability i.e. a declining population, increasing proportion of seniors, and increasing number of household seniors (Yamazaki et al., 2012). Japan, Italy, Germany and Sweden represent 20.1%, 19.7%, 18.8% and 17.2% of people above the age of 65 respectively (Yamazaki et al., 2012). Increased competition, and high wage rates is just another reason for desire for automation (Surana et al., 2020).

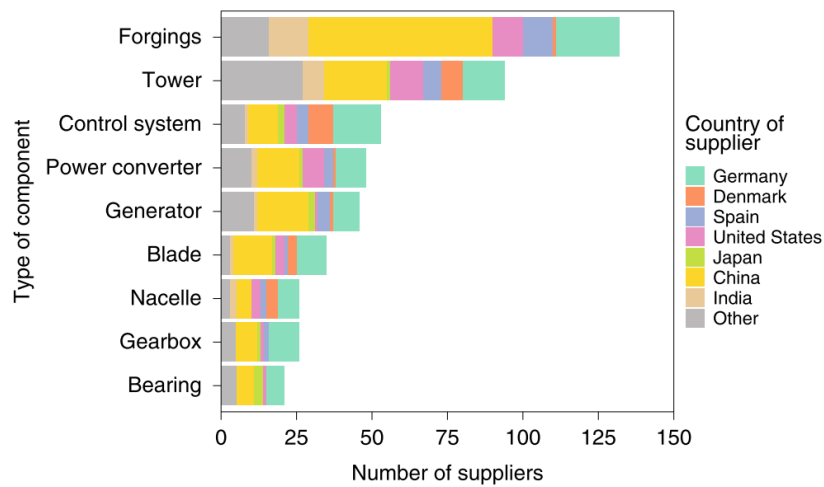

Figure 2: Diversity in number and geographic spread of suppliers by wind turbine component between 2006 and 2016 (Surana et al., 2020).

The manufacturing of wind turbines is largely comprising of manual processes. But due to the large sized components, several assembly tasks are becoming challenging to be performed by humans. Additionally, since wind turbines are at an early phase of their design evolution, frequent design variations do occur due to which fixed automation solutions are not feasible (Perzylo et al., 2019).

Collaborative robots can operate in close proximity to humans and can enable flexible automation enabling a scenario to combine the best skills of machines and humans (Malik et al., 2019) (Lv et al., 2021). Collaborative robots is an innovation in industrial robots offering features to be designed as safe for humans, mobile, and relatively easy to reconfigure to new situations (Malik & Brem, 2021). Cobots combined with safety technologies (Malik & Bilberg, 2019) and with the ability of being mobile and interactive can be used to form human-robot team (Malik et al., 2021) in addressing the challenges of wind turbines manufacturing.

This paper presents:

5. A systematic framework to develop a mobile robot assistant in wind turbines manufacturing
6. A case of developing a robot assistant to assist operators in nacelle assembly
7. Developing a digital twin to speed up the validation and development of robotic assistant
8. Development of HRC interaction strategies for industrial applications
9. Future research directions are discussed

## 2. Related work

Manufacturing automation brings safety, efficiency, comfort and cost effectiveness (Boy, 2018). Industrial automation of physical tasks is largely attributed to large and bulky industrial robots (Kurfess, 2004) that are immobile, operate in fencing and are time consuming to reconfigure (Pedersen et al., 2016). They are classified as fixed automation solutions offering high production volume, strict separation from humans and little to no flexibility (Ji et al., 2021).

Humans can adopt to changing scenarios (Parasuraman & Wickens, 2008) and if are merged with automation solutions can form flexible automation. This idea emerged the concept of human-robot collaboration in manufacturing settings (Wannasuphoprasit et al., 1998). Collaborative robots (or cobots) have created automation opportunities of tasks which were not possible to automate with conventional robotics (Malik & Brem, 2021) (Simões et al., 2022). Cobots are being used in automation of tasks such as pick and place, assembly, welding, inspection, and packing (Ajoudani et al., 2018). Cobots have also been explored as a solution for repurposing of factories in emergency situations (Malik et al., 2020). However, most cobot applications have been in small components manufacturing such as electronics, appliances, and actuators (Malik et al., 2019).

Mobile robots are another form of collaborative robots (Bänziger et al., 2017). These robots find their use in a wide range of applications such as factories, military operations, healthcare, search and rescue, security guards and homes (Shneier & Bostelman, 2015). They are suitable for applications requiring frequent interactions with humans. In manufacturing settings, the trend is towards using mobile robots for material handling applications (Shneier & Bostelman, 2015).

A combination of an articulated robotic arm and a mobile robot often is referred to as a robot assistant (Hägele et al., 2002). A review of mobile assistive robots in automotive manufacturing and assembly has been presented by (Kirova, 2020). The study also presented the requirements of a mobile assistant in automotive manufacturing. A home assistant robot for aging population was discussed by (Yamazaki et al., 2012) (see Figure 3).

Table 1: General criteria for an industrial use of assistive robotics in automotive industry (Kirova, 2020).

| Industrial requirements for assistive robotics | |
|---|---|
| Navigation | Robustness in unstructured environment |
| Gripping technology | Applicability for different part geometries |
| Hardware components | Economic components with compliance of industrial standards |
| Workload | 20 kg 1.8 m |
| Workspace | 99 % 24 hours |
| Availability | CE |
| Energy supply | 20 kg 1.8 m |
| Safety | CE labelled application for man-machine interaction |

Aircraft manufacturing has similarities with wind turbines manufacturing. The Air-Cobot project (Donadio, F., Frejaville, J., Larnier, S. and Vetault, 2016) reports the development of collaborative mobile robotic assistant for airport based routine inspection of aircrafts before takeoff. Robotic assistant in aircraft assembly has been reported by (Kheddar et al., 2019). However, the work is about humanoid robotic assistant with bipedal walking.

The application of human-robot collaboration in large equipment manufacturing is limited in literature. The reason is the limitation of pay load capacity, and the safety of co-existing humans. It is also desired for the robots to be flexible to adapt to frequent design changes.

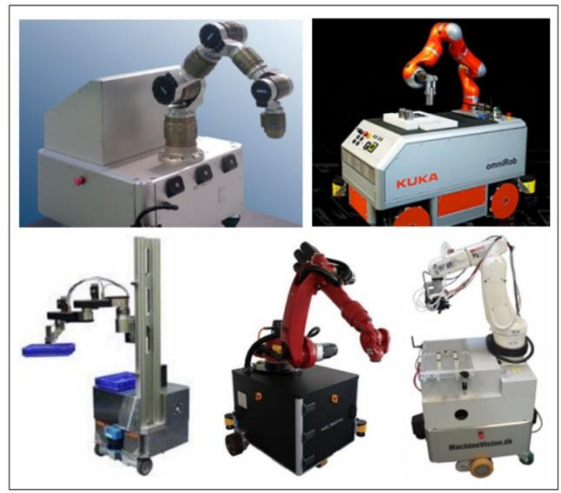

Figure 3: Mobile assistants for industrial use (Kirova, 2020).

## 3. Development of a robot assistant as a co-worker

A mobile robotic assistant refers to a device that moves from place to place to achieve a set of goals (Shneier & Bostelman, 2015). If used as a coworker, it needs to imitate a range of skills retained by humans such as receiving information, processing information, and accomplishing a task to effectively cooperate with the fellow human(s). In the proposed method, the development of robotic assistant consists of four phases (Figure 4). The process starts with a problem description, and an initial idea and concept where the objectives of developing a robot assistant are identified, as well as the financial paybacks are evaluated. The next phase is to design the robot system, that includes selecting (from off the shelf solutions) or developing a robot manipulator and associated hardware. To enable flexibility, a tool changing mechanism can be integrated thus allowing the robot to perform a variety of tasks.

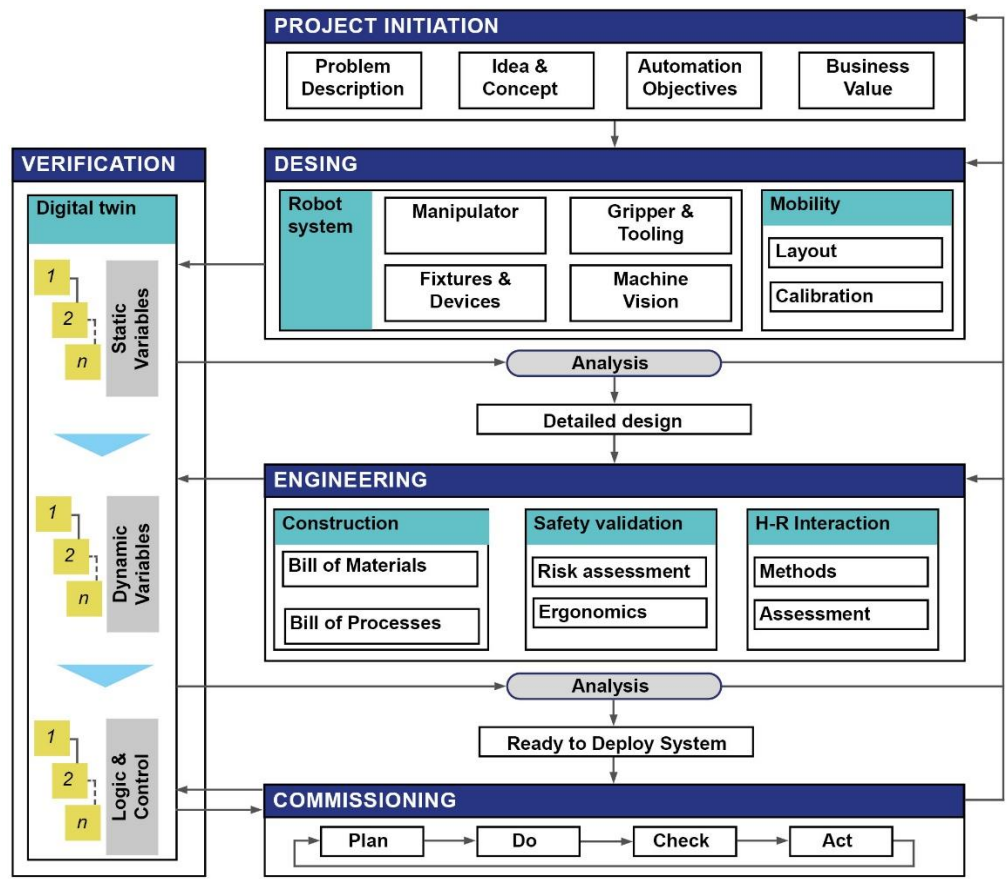

Figure 4: Generic process flow of developing robotic assistant.

The design data is sent to a simulation based digital twin. The results are evaluated against the objectives and business value, and a detailed design is achieved. The next phase is to develop or to construct the system as per locked design. Safety risk assessments are evaluated, and all this information is again fed to the digital twin. Several form of logic and control simulation can be developed. Cycle times are estimated, robot codes are generated and finally a robot system is conceived. The next phase is to implement the robot assistant in serial production.

The development process is described with the help of a use case.

## 4. An industrial case

A case from a leading wind turbines manufacturer is presented to demonstrate the design and development of a mobile robot assistant. The case presents the serial production of >10 MW offshore wind turbines.

## 4.1. Problem description

A wind turbine is primarily consisting of three components i.e., nacelle, blades, and tower. Bed frame is an important part of nacelle and holds several components. It also facilitates the rotary movement of rotor to compensate the change in wind direction. A network of electric cables is mounted both inside and outside of bedframes. A series of cable trays are assembled on the surface of a bed frame to hold these electric cables (Figure 5). There are seventeen cable trays assembled in the inside and thirty-six on the outside of a bed frame in the use case example. These cable trays are assembled using magnetic clamps which are part of cable trays. Accurate placement of cable trays is important to ensure correct alignment of electric cables. An accuracy of ± 5mm is required for assembly of cable trays on bedframes.

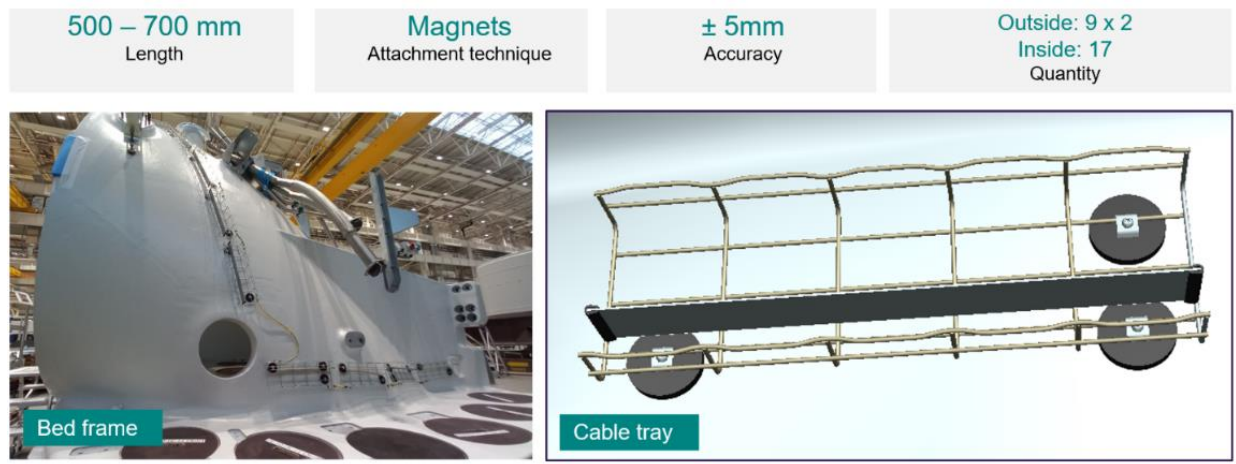

Figure 5: Details of cable trays and their placement on bed frames.

The cable trays are assembled manually in the current practice. Manual assembly is done by a human operator standing on an elevating work platform (EWP). Though technical drawings are available illustrating the placement location of each cable tray but making linear manual measurements is difficult due to large and curved surface of the bed frame. Therefore, the operator assembles the cable trays on the curved bedframe surface by manually observing an optimal location (see Figure 5).

The manual assembly brings following challenges:

- Making manual measurements is a time consuming and repetitive process.
- Since the operator works close to the bedframe, the depth perception makes it difficult to remain aligned with the reference points during assembly.
- Most cable trays are interconnected via cables-wires complicating the handling of cable trays. Minor variations due to misplacements add up resulting into large dimensional errors.

The resulting assemblies can have a large degree of variation (up to ± 100mm). When the electric cables are assembled in the cable trays, they can result in to too short cable lengths. Such a situation will require repositioning and realignment of cable trays. The repositioning can induce several hours of re-work and even production halts can happen until the error is resolved.

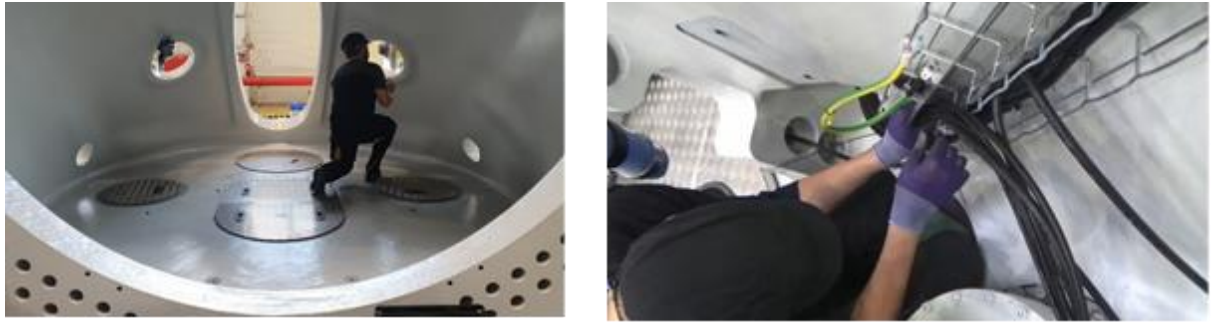

Figure 6: Comparison of cable trays placements of two different bed frames after manual assembly.

### 4.2. Idea and concept

A robotic assistant is proposed to facilitate the operator by highlighting the target assembly locations and eliminating the need of making any manual measurements. The robot assistant uses a laser device at its tool post to project laser markings onto the bedframe's surface for each cable tray location. The robot points at each assembly location until all the cable trays are assembled. The robot assistant is desired to bring accuracy, repeatability and relieve operator from repetitive task (see Figure 6).

Table 1: Desires features of the robot assistant.

| Feature | Must have | Nice to have |
|---|---|---|
| Laser beam | Create a visible beam of laser at the bed frame | The shape of the laser must be editable |
| | Laser beam must define not only the location but also the orientation of the cable tray | Laser beams must be thin and sharp lines |
| Assembly locations | Must offer an assembly accuracy of within 2 to 5 mm | Reconfiguration due to design changes must be fast |
| Human-robot interaction | The operator can interact with the robot from a distance | The operator should be able to interact with robot in natural language |
| Robot localization | The robot must localize itself | Vision based localization must be enabled |

### 4.3. Design of the robot assistant

The robot assistant primarily consists of a robot manipulator, a mobile base, and a laser tool (Figure 7 & 8). A description of each of them is given below:

**Robot manipulator:** The robot manipulator used is Doosan M0609. It is a collaborative robot, has 6 degrees of freedom (DoF), reach of 900mm and payload capacity of 6 kg. The robot has a cockpit at its arm that can be used for quick interface to the operator.

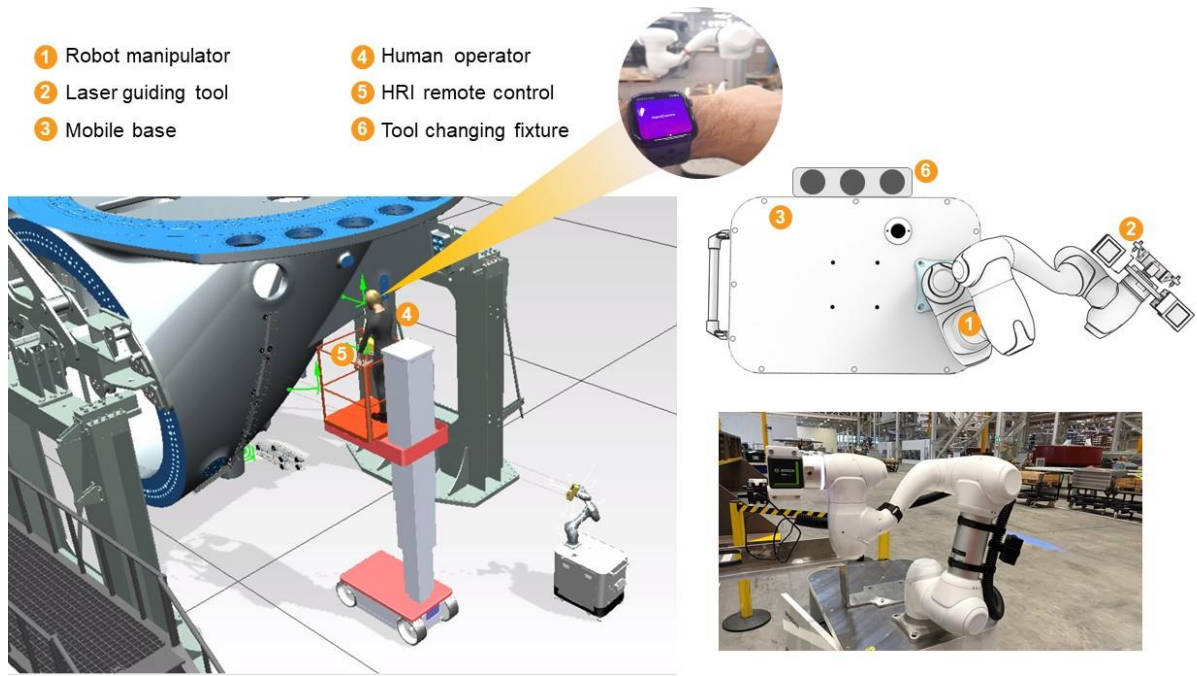

Figure 7: Design of the robot assistant.

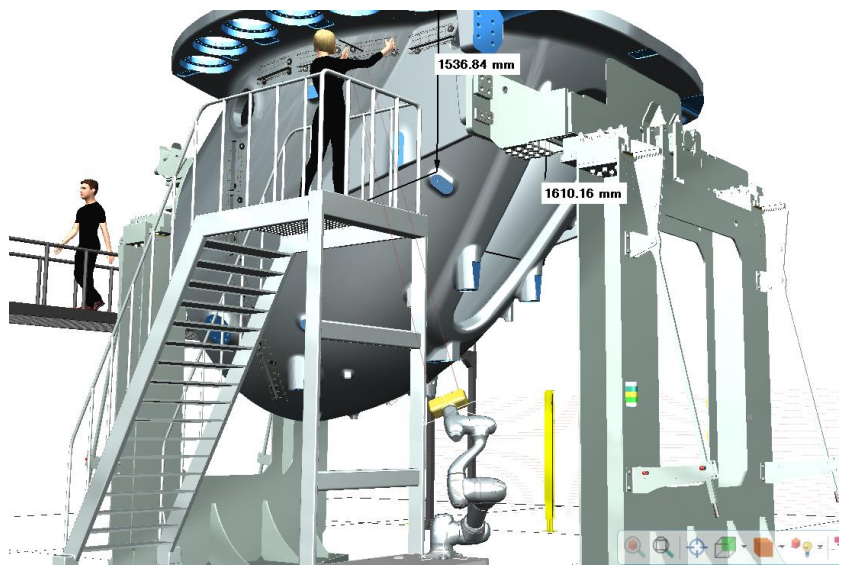

Figure 8: Application of robot assistant in wind turbines manufacturing.

**Mobile base:** The robot manipulator is mounted on top of a mobile base. The mobile base allows the robot for easy movement and maneuvering of the robot. It also has storage available for tools and peripherals. The dimensions of the base are 850(W) X 600(D) X 825(H) mm and has a weight of 166 kg. The top surface of the mobile base is made of stainless steel. For the prototyping purposes a manually portable mobile base is used, but it is highly recommended to use an autonomous mobile feature for future tests.

**Robot tooling:** The robot tool is desired to be flexible, and cost-effective. A flexible tool will allow to perform more than one tasks (to hold a variety of tools/devices) and accommodate future variations in product design. A modular tool was designed and produced using additive manufacturing (see Figures 9). The tool can hold four different lasers. Furthermore, a calibration camera can also be mounted on the tool.

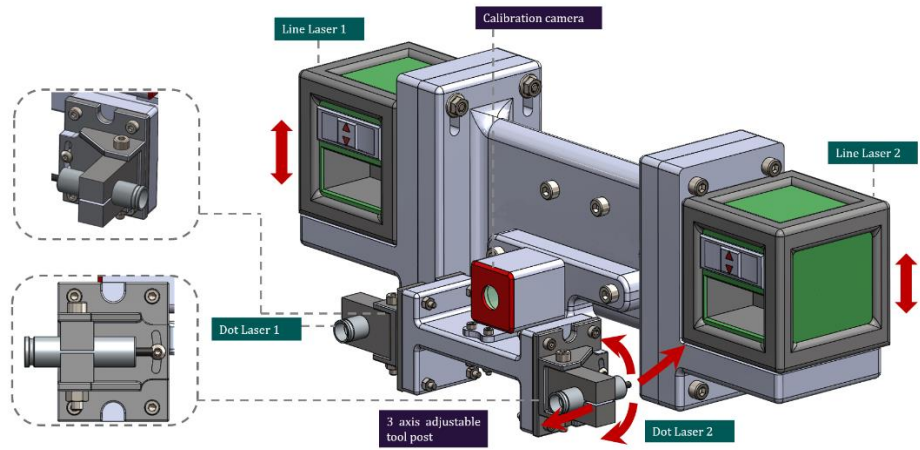

Figure 9: Design of the laser tool for the robot assistant.

For simplified fastening of the tool to the robot, a snap mechanism (for quick attach and detach) of the laser tool is enabled. As a result, the tool can be attached or detached using snap fit and no other fastener is needed.

### 4.4. Laser markings to accurately point at assembly locations

The system required a laser device that can reasonably project laser beam on the bed frames. The laser pointing must be visible at a distance of 8 – 10 meters to guide the operator of assembly locations. The BOSCH Quigo Laser was selected for the purpose. It is a cost-effective laser device generally used as a laser leveling tool for maintenance tasks at home. A set of two laser devices were attached to the robot tool (see Figures 10).

Every laser possesses an error in its laser beam, meaning a laser beam moves at an angle <0. To project the laser beam at the right locations of the cable trays according, the error needs to be compensated in the robot program. The error is calculated using the real lens which is difference of where the laser beam should project and the where is projects in real. The difference is modelled into the simulation to account for the error.

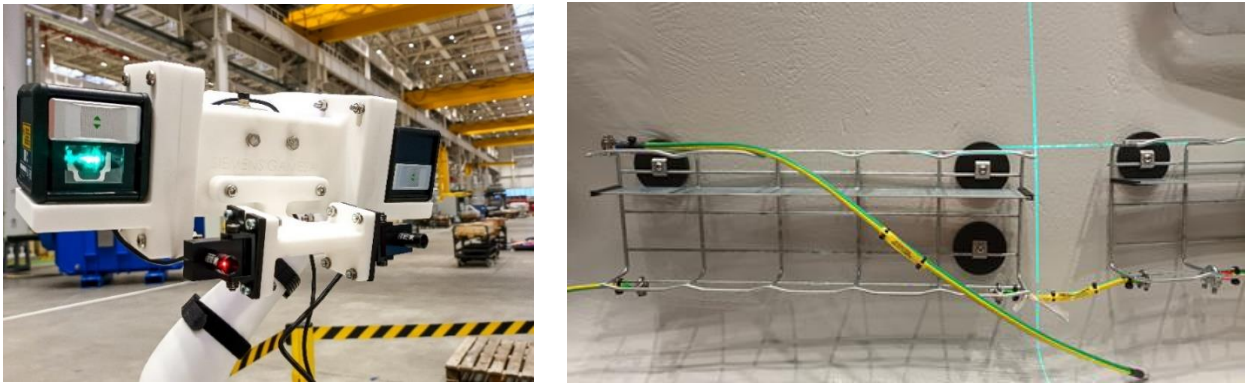

Figure 10: Developed laser tool with additive manufacturing.

### 4.5. Layout design

The robot needs to maneuver around the workspace to properly guide the worker for assembly locations (thanks to the mobility feature of the robot platform). It is proposed, considering the laser range and robot reach, that the robot maneuvers to several locations in the workspace to perform the task. Five different locations are identified in the workspace for the robot to perform its assigned tasks. The layout of the workspace is shown in Figure 11.

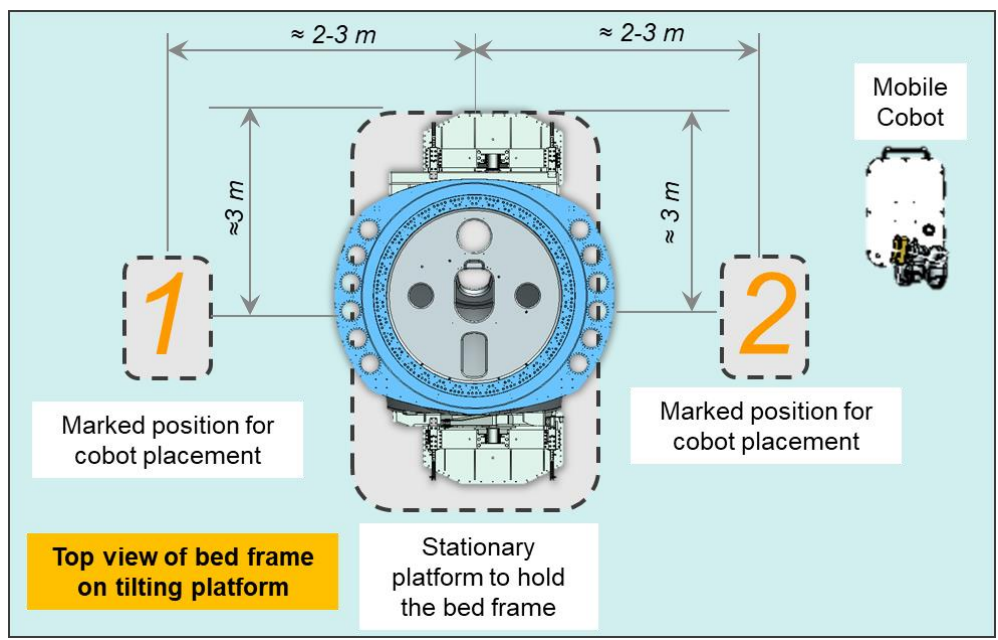

Figure 11: Robot workspace.

### 4.6. Localization of the robot system

Localization or calibration of the robot assistant is needed to make the robot "know its world". A right localization technique must offer accuracy, repeatability, cost effectiveness and reconfigurability. It must also adopt to the assembly locations as it maneuvers to different locations or in other words, the robot must be able to determine its own location with respect to the bed frame. The robot system needs to be localized both for the internal cable trays as well as for the outer cable trays (Figure 12).

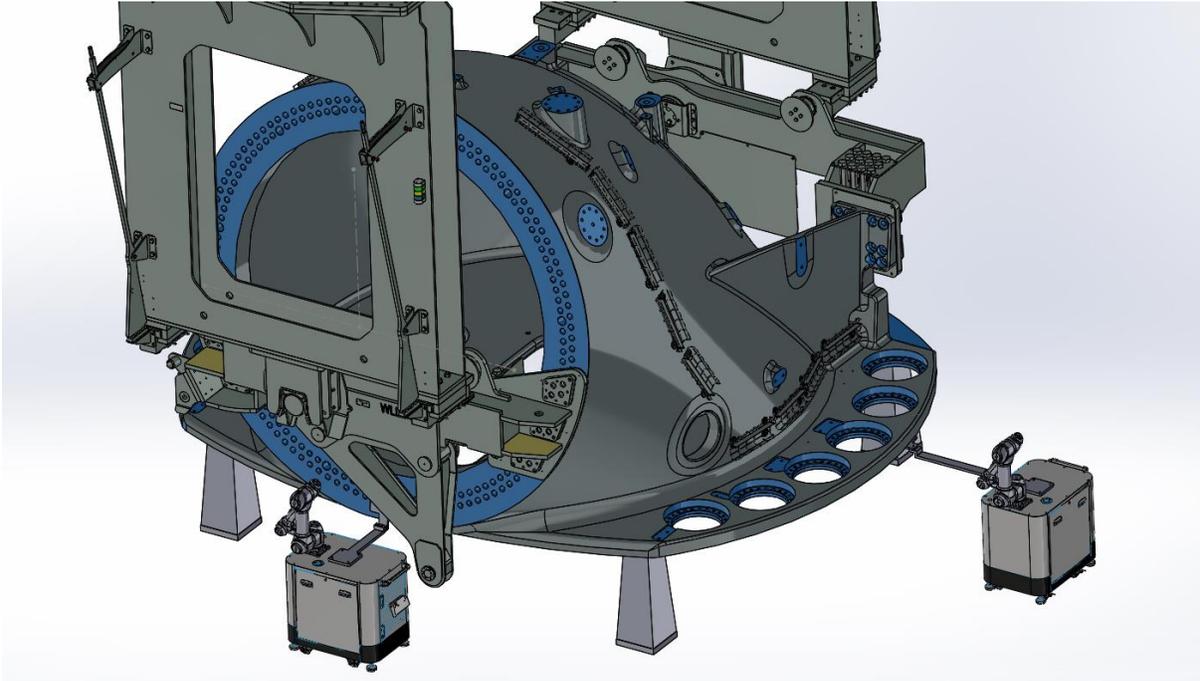

Figure 12: Robot localization with mechanical fixtures.

Initially, a mechanical fixture-based localization method was adopted. The mechanical fixtures are aligned with fixturing features in the bed frame (Figure 13). The fixturing features on the bed frames remain constant and hence reportability can be achieved. However, an intelligent and automated system requires to have vision-based localization. There are different techniques for vision-based localization such as markers, 3D cameras, and laser-based localization.

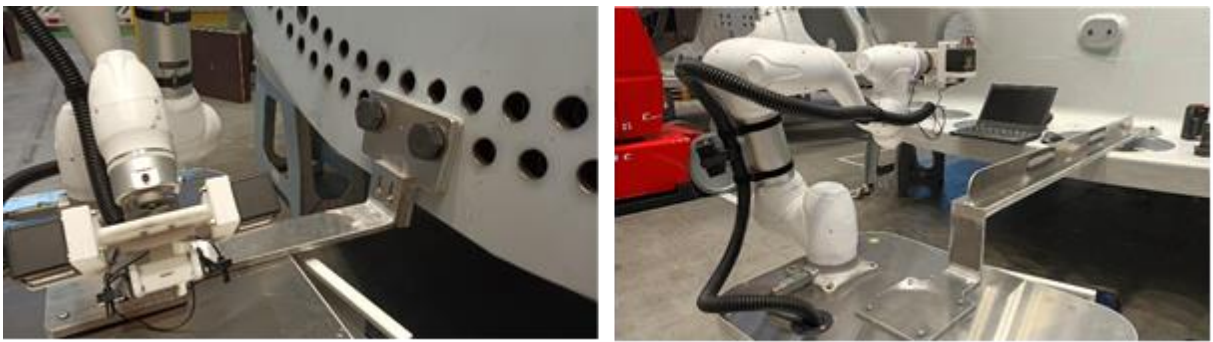

Figure 13: Robot system localization. Fixture for localization on front of the bedframe (left); fixture to be placed on the side of the bed frame (right).

### 4.7. Digital twin to design, validate and commission the robot assistant

A digital twin is a virtual copy of the physical system that supports the design, development and commissioning of the robot assistant. Several types of validation tests can be made in a comprehensive digital twin to speed up the validation process (Figure 14).

Some of the benefits of using a digital twin are:

- It was an objective to use the CAD models of bed frames as reference data. It was a goal to extract cable trays' positions from the CAD models and transfer them to the bed frames. This will ensure that the already approved design of the bed frame and cable trays is being complied. Therefore, a complete workspace with a robot assistant was modelled in the simulation during the design phase. It accurately presents a dynamic simulation of the process. The modelling of laser beam was a challenging task because the robotic simulation software is not capable to model laser lights or beams.

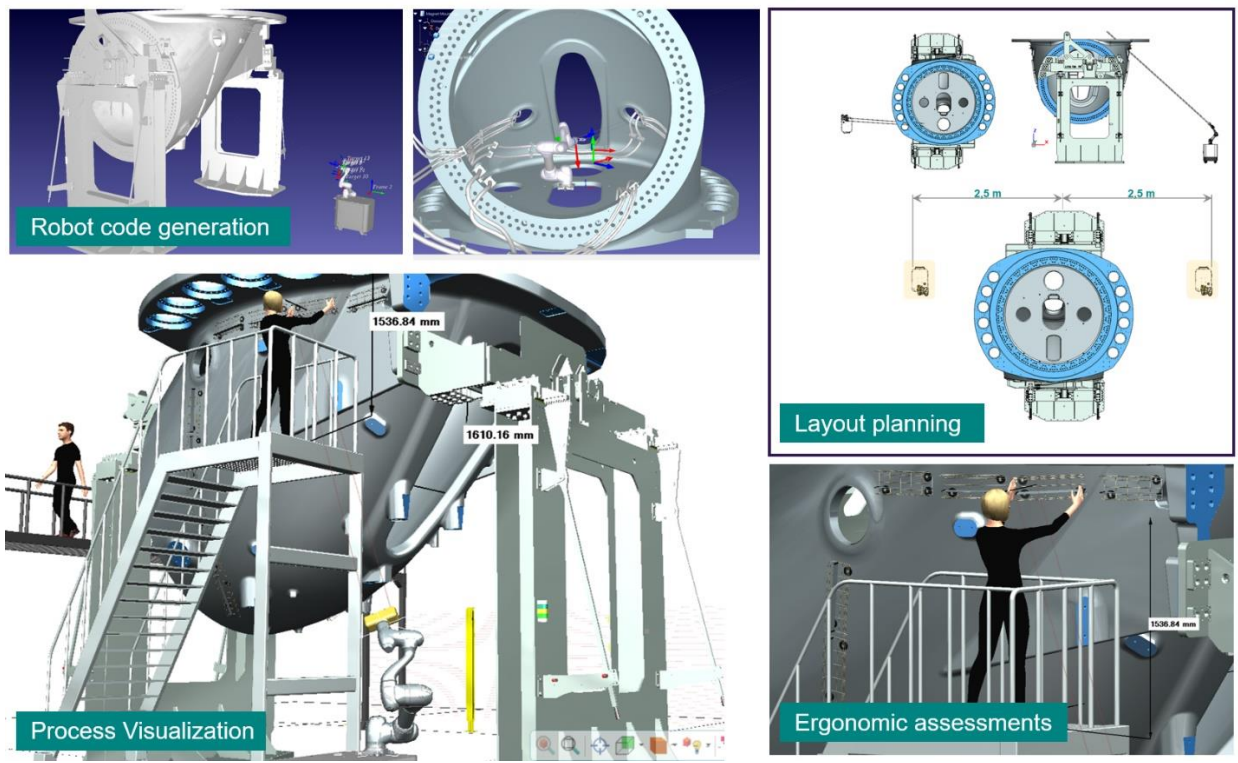

Figure 14: A digital twin of the robot assistant for online programming.

- Another challenge was that the laser beam (from the laser used in the project) is not a straight line but has an offset. This offset was measured manually from the laser and was modelled accordingly in the simulation. All the CAD data was saved in STEP (Standard for the Exchange of Product Data) format, that is acceptable by RoboDK. After this the CAD models were imported into the RoboDK software.
- The simulation is live connected with the robot using an ethernet connection. For this purpose, the I.P. address of the robot can be obtained from its settings and is used in the Connection settings of the RoboDK software. RoboDK must 'know' which robot brand is being connected to it (to optimize the robot program to that specific robot language). A post processor in RoboDK is available for Doosan Robotics and must be selected. A robot post processor defines how robot programs must be generated for a specific robot controller.

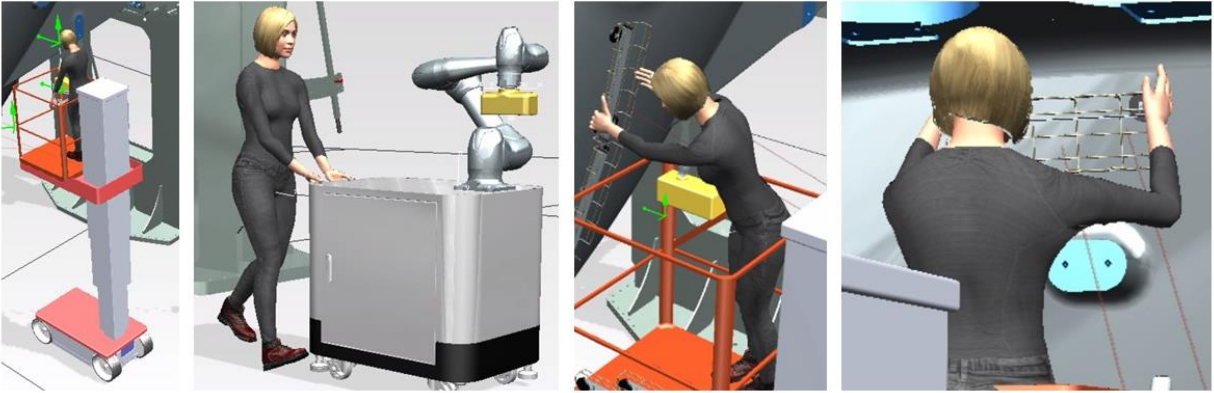

Figure 15: A digital twin of the robot assistant for online programming.

- Once the robot is connected to the computer through a TCP/IP port, the IP address of the robot is entered in the Robot IP tab. And is connected with the robot. Once the connection is established, the robot gets disconnected from it teach pendant and control is transferred to RoboDK. In this environment, the robot positions from the simulation can be transferred to the robot (using **Move *Joints***), and robot positions from the real robot can be transferred back to the simulation (using ***Get Position***).
- Detailed human simulation is developed for the whole process and for most human-robot interactions (Figure 15). Using collision detection, collision avoidance, and work envelop tests, the layout and positioning of equipment was modified.

## 4.8. Build

This step involves development and commissioning of the robot assistant. Hardware in the loop (HiL) simulations are developed to verify logic driven simulation. The digital twin helped to perform most of the tests in the virtual environment and once it reaches to a satisfactory maturity then it is translated into real world.

## 4.9. Test results

The tests showed an overall compliance with the desired objectives, namely the projection of CAD-associated cable tray locations and orientations onto the real bedframe. The laser beam was visible on the bed frame and locations and angles could be clearly identified by the laser with satisfactory accuracy (Figure 16).

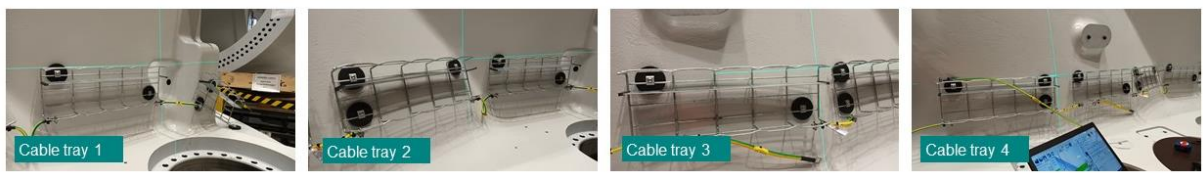

Figure 16: Results from the test.

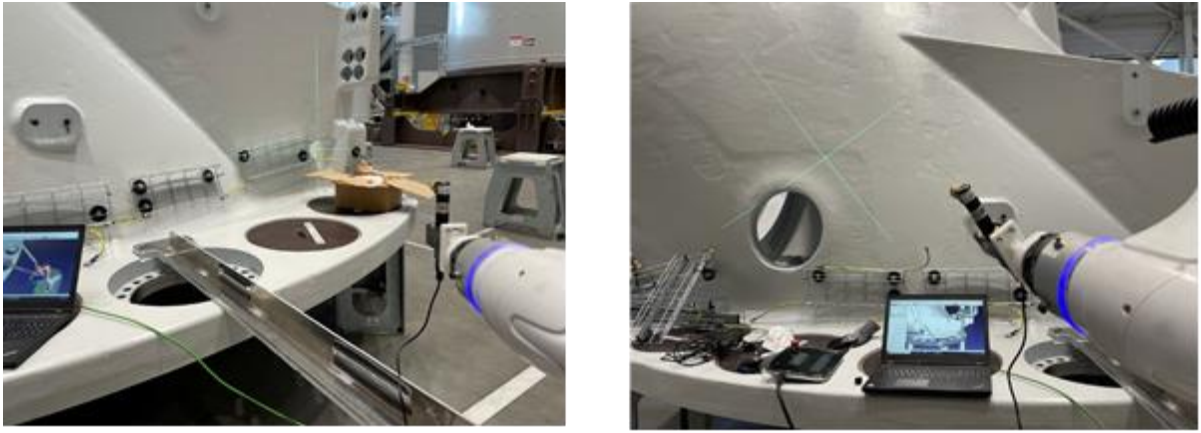

Figure 17: Results from the test.

During mounting of the vertical cable trays, the ground cables that are pre-assembled to the cable trays, were mostly too long and thus produced too tight angles. For correct mounting, they need to be rather straight. The operators then opened the ground cable clamps and repositioned them (Figure 17 & 18).

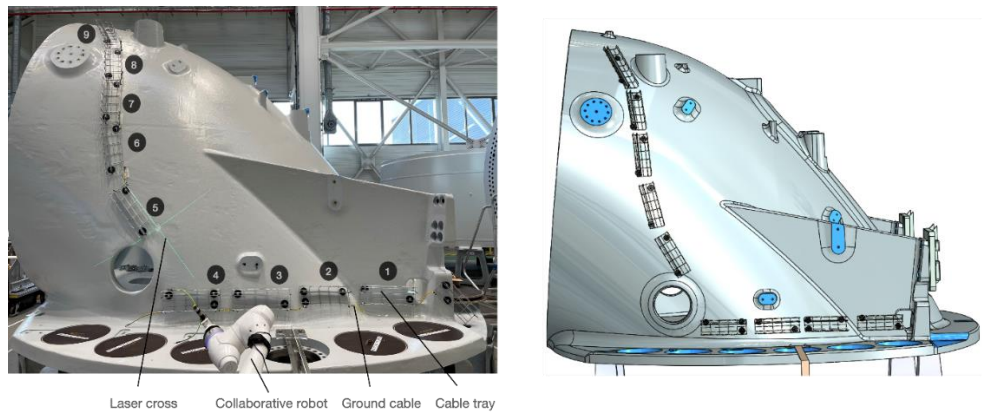

Figure 18: Results from the test.

### 4.10.        Operator to robot interaction

The operator needs to fluidly interact with the robot during tasks execution to command the robot to move to next/previous tasks, or to restart the process or to come to a complete stop. A wireless interface is required to enable the operator to freely move around in the workspace. The interface also needs to be simple and user friendly. Different HRI techniques or industrial application are shown in Figure 19.

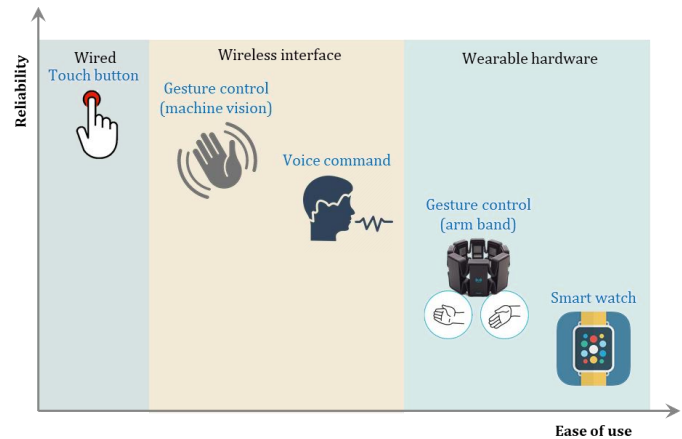

Figure 19: Various techniques for human-robot interaction.

To provide a simple and wireless interface prototypes of HRIs were developed including smart watch, remote control and smart phone (Figure 20). Given the cost and ease of implementation, remote control was found most feasible. The remote control was interfaced with a smart switch (Sonoff 4CH Pro). The Sonoff switch has four channels and each of the four buttons of the remote control can be interfaced with them. Schematic diagram of the implemented remote control is shown in the Figure 21.

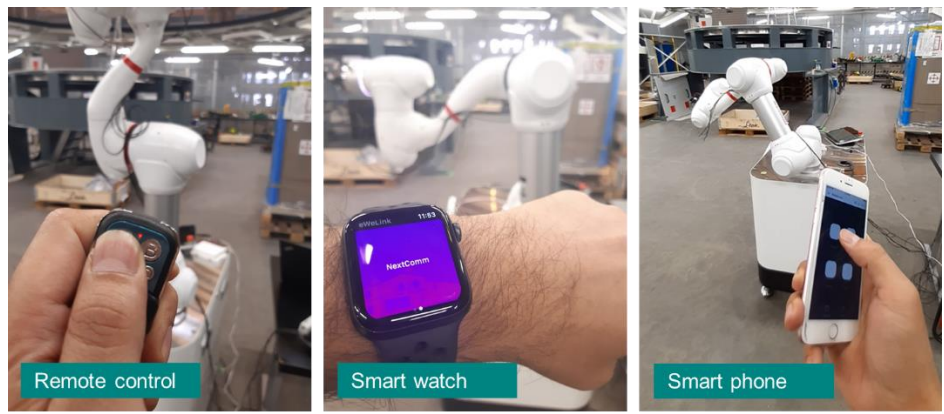

Figure 20: Developed prototypes for HRI interaction.

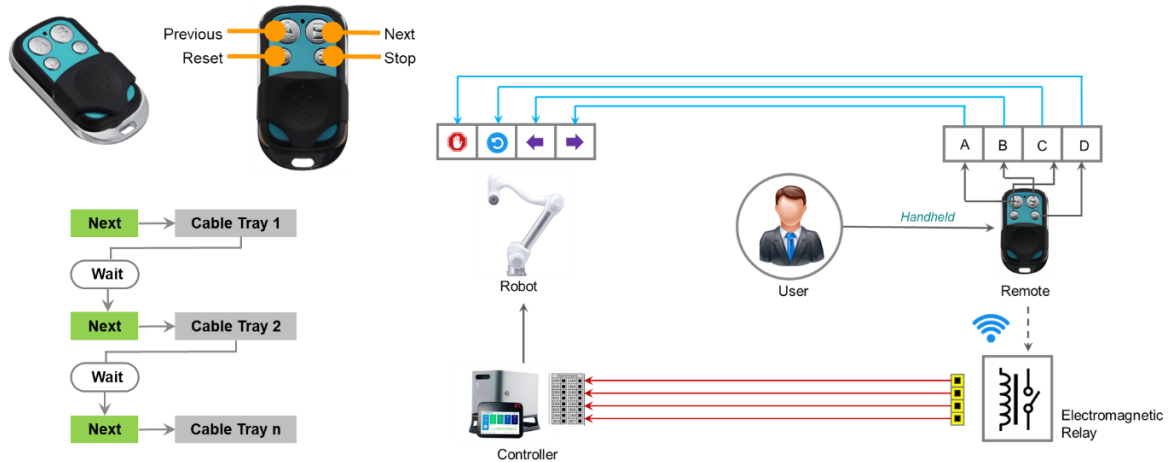

Figure 21: Schematic of interfacing the remote control with robot controller and robot program.

## 5. Limitations & further work

The limitations of the project and further work are presented below:

**Robot calibration:** The calibration or localization of the robot is a fundamental part of the project. In the current setup manual localization technique is used with mechanical fixtures, however, the repeated use of mechanical fixtures is not advisable due to their large size, weight and potential harm to the robot system (e.g., damaging robot platform surface or accidentally hitting the robot). Therefore, a vision-based calibration system will make it more convenient and durable.

**Alignment with safety standard:** Care has been taken to design the system in a way that it complies with the safety limitation defined in ISO-15066 safety standard. For example, the safety monitored stop, force limiting body of the robot, and avoiding any sharp edges in robot tooling etc. However, for any change in the physical workspace a thorough safety risk assessment will be required for the robot, the robot system and its application.

**Mobility:** The robot assistant is planned and researched as an autonomous mobile robot. However, for practical reason, the developed prototype is not comprising the feature of autonomous mobility. There are several mobile robots available in the market that can be used for this purpose. However, the future tests must validate the scenario of autonomous mobility.

**Reducing overall cost of wind energy:** Robotics and particularly collaborative robots have potential in large scale manufacturing to reduce person hours and increase productivity. More studies are needed to identify other applications of cobots in wind turbines manufacturing and assembly. It also needs to be identified if cobot technology can result in reducing the cost of energy.

## 6. Conclusion

The visual cues provided by the robot assistant for the operators during assembly tasks have not only simplified the assembly process but also reduced the assembly time. The operators found the robotic assistant useful and felt comfortable to work alongside it. Nevertheless, the challenges identified in its application were the cost of safety technologies, reconfigurability, localization of mobile robot, and enabling an intuitive/user friendly HRI. Additionally, since wind turbine manufacturing is a semi-fixed layout production, there is idle time available when the robot is not being used. Therefore, the cobot is designed to be reconfigurable for additional tasks. It can be done by changing the end of arm tooling and downloading a new robot program. An example is the assembling of another component (magnet mountings) on bed frames. Several magnet mountings are used in a bed frame to hold the cooling hoses. The assembly of these magnet mountings has also been facing the same challenge. With little modifications in the laser tool, the robot is equally useful for magnet mountings assembly. Though there are limitations to what tasks cobots can take over. But the article argues that the low payload capacity of cobots should not expel them from applications in large components manufacturing. The cost of manufacturing of wind turbines needs to be reduced and cobots can play a role by increasing the level of automation. Reducing the cost of wind energy and promoting its accessibility will eventually play an integral role in achieving all the MDGs of UNO.